# Towards Practical Application of Deep Learning in Diagnosis of Alzheimer's Disease.

Harshit Parmar, Eric Walden

Texas Tech Neuroimaging Institute, Texas Tech University, USA

**Abstract**

Accurate diagnosis of Alzheimer's disease (AD) is both challenging and time consuming. With a systematic approach for early detection and diagnosis of AD, steps can be taken towards the treatment and prevention of the disease. This study explores the practical application of deep learning models for diagnosis of AD. Due to computational complexity, large training times and limited availability of labelled dataset, a 3D full brain CNN (convolutional neural network) is not commonly used, and researchers often prefer 2D CNN variants. In this study, full brain 3D version of well-known 2D CNNs were designed, trained and tested for diagnosis of various stages of AD. Deep learning approach shows good performance in differentiating various stages of AD for more than 1500 full brain volumes. Along with classification, the deep learning model is capable of extracting features which are key in differentiating the various categories. The extracted features align with meaningful anatomical landmarks, that are currently considered important in identification of AD by experts. An ensemble of all the algorithm was also tested and the performance of the ensemble algorithm was superior to any individual algorithm, further improving diagnosis ability. The 3D versions of the trained CNNs and their ensemble have the potential to be incorporated in software packages that can be used by physicians/radiologists to assist them in better diagnosis of AD.

# INTRODUCTION

Deep learning is a type of artificial intelligence algorithm which is capable of extracting useful features from the data without the use of dedicated feature extraction tools. Deep learning techniques are applied directly on unmodified data types. Deep learning is finding new applications in various domains like self-driving cars, natural language processing, computer vision etc. [Goodfellow et al., 2016]. This study focuses on practical application of deep learning techniques for early detection of Alzheimer's Disease (AD). Deep learning techniques are widely used for images. In this study, 2-dimensional deep learning techniques (which are commonly used for images) are modified to 3-dimensions to make it suitable for direct application to neuroimaging data.

Brains are 3D structures. Using a 2D model for analysis of brain structures may cause some loss of information. Figure 1 below shows the comparison of 2D slices and 3D volume for MNI brain atlas. Figure 1a shows three orthogonal slices along the axial, sagittal and coronal planes. No matter what plane is chosen, a 2D algorithm can only extract information from that plane itself.

No information about adjacent slices is obtained. For neuroimaging data, such an in-planar only approach may not be sufficient as useful features may span across all 3 dimensions calling for a 3D algorithm. A 3D brain atlas is shown in figure 1b, and it obvious that a local neighborhood information in the all the directions can be obtained from a 3D volume. Using 3D models preserves special information but practical application of 3D deep learning models is challenging due to availability of limited dataset, increased computational complexity and increased training time. As discussed later in this section, various approaches have been used to indirectly incorporate the 3D information with deep learning. Here we propose the use of a direct 3D deep learning approach for feature extraction and classification of various stages of AD using minimally preprocessed full brain 3D anatomical MRI data.

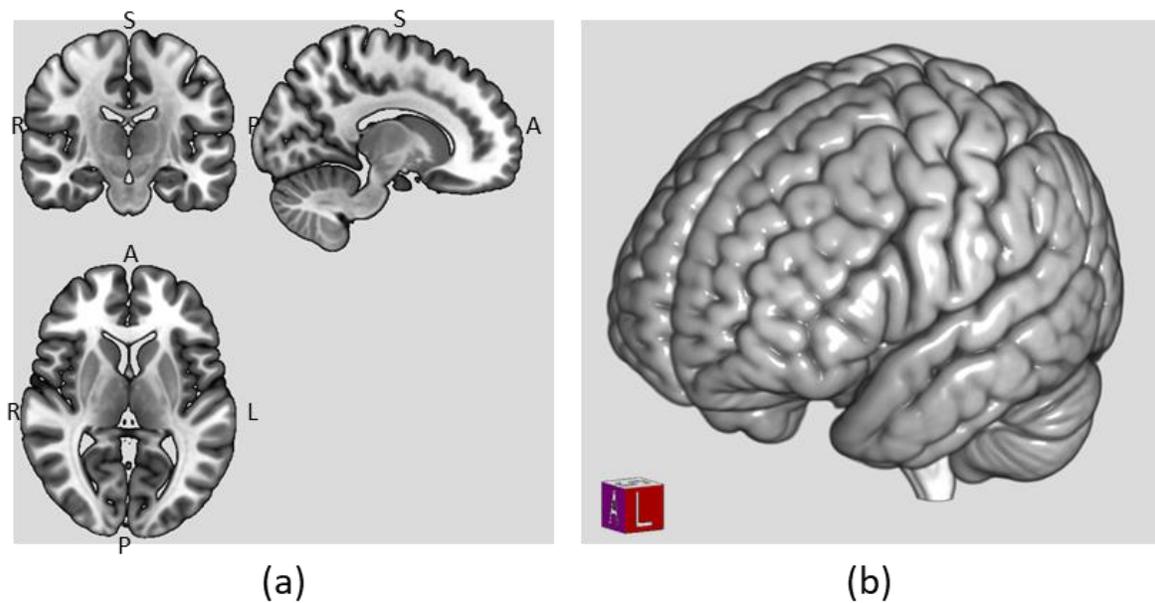

*Figure 1: Comparison of (a) 2D slices and (b) 3D volume for MNI brain atlas.*

Dementia is a brain disorder in which there is disturbance of multiple higher cortical functions including memory, thinking, orientation, comprehension, language, judgement etc. Dementia is mainly observed in elderly people. Almost 98% of the cases reported are of the age 65 or above [WHO, 2006]. The prevalence doubles every 5 years after 65 years and is one of the major causes of disability in later life. AD is one of the most common forms of dementia accounting for 50% to 75% of the total cases of dementia. AD is characterized by cortical amyloid plaques and neurofibrillary tangles with the brain.

Currently there is no cure for AD. There is however evidence that some drugs like cholinesterase inhibitors can reduce the rate of cognitive decline [Geokoop et a., 2004; Goekoop et al, 2006; Miettinen et al, 2015]. Moreover, the symptoms of early onset of the disease are often misinterpreted as normal aging and by the time the disease is diagnosed, it has already reached a later stage. Because of that, detection of the disease in the earlier stage is desirable and an open challenge. If AD is detected in earlier stages, the rate of cognitive decline can be reduced by proper medication leading to a better life for the patient.

Current research suggest that deep learning can be a tool for better and effective diagnosis of dementia using different medical imaging modalities. According to PubMed records, around 700 papers have been published with application of deep learning for AD diagnosis in the last decade [https://pubmed.ncbi.nlm.nih.gov/?term=%28alzheimer%27s+disease%29+AND+%28deep+learning%29&filter=years.2012-2022]. Out of those 700, about 500 research papers have been published in the last 3 years only. This explosion of publications suggest promising potential of deep learning algorithms in diagnosis and classification of AD. A complete comprehensive review of all the deep learning literature with application to AD is out of the scope for this article. However, some of the different deep learning approaches for AD are discussed later in this section.

One of the most flexible and versatile deep learning algorithms is the Convolutional Neural Networks (CNN). Presently, CNNs are considered to be the state-of-the-art machine learning algorithm for classification, segmentation and generation of images [Yamashita et al., 2018; Li et al., 2021]. Over the past decades, many CNN architectures have been developed for specific image processing applications. Some of the earliest application of CNN dates back to 1995 where it was used for identification of handwritten digits [LeCun et al., 1995]. With the advancement of computer hardware and graphics processors (GPUs), CNNs can now perform much complicated. However, the majority of the CNN development is centered around images (2-dimensional data). To make use of CNNs for volumetric data, especially medical imaging scans, modifications have to be made to the data and/or the CNN to make them compatible with one another.

Some of the early works for classification of AD uses 2D CNN on anatomical MRI images. All or specific slices from the entire brain volume are used with 2D CNN for feature extraction and classification. When using multiple slices as input, each slice is considered as a separate data point. Researchers have tried using all axial slices [Sarraf & Tofighi, 2016], a few axial slices depending on the entropy information of each slice [Jain et al., 2019] or single slice overlapping with the hippocampus region [Wang et al., 2018]. Along with axial slices, CNN have also been trained using coronal [Bae et al., 2020] and sagittal [Puente-Castro et al., 2020] image slices as inputs. Along with anatomical images, researchers have also used processed images to extract features and then use the features with CNN to make predictions [Hett et al., 2018].

When using a 2D CNN, some of the spatial information is lost. The use of a 3D CNN allows for the use this spatial information, but with significant added computational cost and complexity. Some researchers have tried to overcome the computation time issue by using a subset 3D volume patch from the entire brain volume [Raju et al., 2020]. A subset 3D volume reduces the input size resulting in faster computation. Some researchers take a step ahead by first using deep learning models to automatically segment the hippocampus area and then use the segmented output with a classification network [Liu et al., 2020]. Such models increase the overall complexity. Another technique which is often used by researchers that increase model complexity is to first train an unsupervised autoencoder (AE) on the 3D volume and then use transfer learning weights from the AE to fine tune a classifier [Hosseini-Asl et al., 2016; Oh et al., 2019].

Apart from MRI, deep learning models have also been used with other modalities like positron emission tomography (PET) [Lu et al., 2018] and magnetoencephalography (MEG) [Lopez-Martin et al., 2020]. Some studies also suggest the use of deep learning on multiple modalities for both classification and extraction of useful features [Huang et al., 2019; Oh et al., 2019]. The application of CNN is not limited to anatomical brain volumes. Deep learning models have also been used with 4D functional MRI (fMRI) data to classify different stages of AD. Multichannel 3D CNNs have been used for both binary [Parmar et al., 2020a] and multiclass [Parmar et al., 2020b] classification of AD using fMRI data. CNNs have also been used with fMRI data to extract useful features from temporal data which in-turn was used by a recurrent neural network (RNN) to predict AD [Li et al., 2020].

There are various deep learning models all of which take advantage of different aspects of neuroimaging data for better diagnosis of AD. Broadly speaking there is an indirect trade off between network training complexity and data preprocessing complexity. For example, a simple CNN may require complex preprocessing and feature extraction before the data is fed into the CNN. Region of Interest (ROI) selection or feature extraction stage is common with simpler CNN. While if raw anatomical brain volume (or part of it) is used as the input, the CNN needs to be relatively complicated to extract useful features. A simple CNN may train faster and require less memory, but the complicated preprocessing may not be practical for every new data point in clinical environment. Keeping that in mind, we believe that it is more advantageous to have a network which takes more time to train but relatively simpler for the actual application purpose. As major portion in training a CNN is a one-time aspect, while prediction on new datapoints is frequent and thus need to be fast. So here we design CNNs which can operate on minimally preprocessed anatomical full brain volumes to make predictions about AD.

We designed and trained 3D versions of three well known CNN architectures and compare their performance for neuroimaging data obtained from the Alzheimer's Disease Neuroimaging Initiative[1]. The three CNN architectures are: Alexnet [Krizhevsky et al., 2012], GoogleNet [Szegedy et al., 2015] and VGG 16 [Simonyan & Zisserman, 2014]. The reason for choosing these three specific CNNs lies in their architecture and popularity. Alexnet is the simplest CNN architecture with series of convolutional and pooling layers for feature extraction followed by fully connected layers for classification. The unique aspect of GoogleNet architecture is the Inception module, consisting of parallel convolutional layers. The VGG16 architecture consists of cascaded (series) convolutional layers between pooling layers. Along with comparison of performance of all three types of CNN (simple, parallel and series convolutions), the feature extraction of all CNNs were investigated. Features that each network considered significant for making a prediction were studied to find similarities with known anatomical features associated with AD. Finally, with practical application in mind, an ensemble framework is presented which improves the diagnostic accuracy for classification of AD. The presented framework has a potential to be developed into a stand-alone software package that can provide additional information along with the traditional brain images.

---

[1] *Data used in preparation of this article were obtained from the Alzheimer's Disease Neuroimaging Initiative (ADNI) database (adni.loni.usc.edu). As such, the investigators within the ADNI contributed to the design and implementation of ADNI and/or provided data but did not participate in analysis or writing of this report. A complete listing of ADNI investigators can be found at:
http://adni.loni.usc.edu/wp-content/uploads/how_to_apply/ADNI_Acknowledgement_List.pdf

# METHODOLOGY

## Dataset and preprocessing

The CNN were trained using anatomical MRI data. Data used in the preparation of this article were obtained from the Alzheimer's Disease Neuroimaging Initiative (ADNI) database (adni.loni.usc.edu). The ADNI was launched in 2003 as a public-private partnership, led by Principal Investigator Michael W. Weiner, MD. The primary goal of ADNI has been to test whether serial magnetic resonance imaging (MRI), positron emission tomography (PET), other biological markers, and clinical and neuropsychological assessment can be combined to measure the progression of mild cognitive impairment (MCI) and early Alzheimer's disease (AD). Anatomical data for 1502 participants were downloaded from the ADNI online database. The participants were from 3 different groups – Alzheimer's Disease (AD), Mild Cognitive Impairment (MCI), and healthy control (CN). The group wise demographic data is shown below.

AD: 225 F / 275 M, age: $75.74 \pm 7.82$ years

MCI: 217 F / 288 M, age: $75.54 \pm 8.12$ years

CN: 213 F / 284 M, age: $75.33 \pm 7.31$ years

All the anatomical data were acquired on a 3T MRI machine with MPRAGE, 3D MPRAGE or accelerated MPRAGE imaging protocol. The online dataset consists of data from different scanning sites and thus the data dimensions are different. As preprocessing, all the volumes were normalized to a standard MNI brain atlas [Fonov et al., 2009] with a spatial resolution of 1 mm x 1 mm x 1 mm and segmented into gray matter, white matter and cerebrospinal fluid (CSF). SPM 12 toolbox was used along with MATLAB 2020a to do the preprocessing [Ashburner et al., 2014]. The segmented volumes acted as 3 channel inputs to the CNNs. Thus, the input size to the 3D CNN was $157 \times 189 \times 156 \times 3$, which is equivalent to 13.8 million datapoints per brain. A total of 1502 brains were used to train, validate and test the CNNs resulting into about 21 billion individual datapoints. Each of the CNN also consists of tens of millions of learnable parameters which were optimized (trained) for thousands of iterations thus making it a big data problem.

## Convolutional Neural Network
### *Architecture*

Some of the most commonly used CNNs are designed for 2 dimensional images while neuroimaging data is typically 3 dimensional. When using a 2D CNN directly on to sections of 3D brain important spatial information is lost which may reduce diagnostic efficiency. Thus, 3D CNNs are more suitable for neuroimaging applications. A major contribution of this study is the implementation and training of 3D CNN. In this study a 3D version of existing 2D CNN is being implemented and trained for classification of 3D full brain MRI data into different stages of AD. Three well known CNN architectures trained in this study are Alexnet, GoogleNet and VGG16. The three CNN architectures were modified to handle 3D data while keeping the base architecture as close to the original 2D image version as possible.

Each of the 3 selected CNN have a unique characteristic in terms of data flow through different layers. In 3D variant, uniqueness of the interconnection of the layers were kept the same. However, changes were made to handle an extra dimension in the data. By simply adding an extra dimension, both time and computational complexity increases significantly. For example, a 3x3 kernel convolution for 64x64 image will require 34596 (62x62x9) multiplications and 3844 (62x62) addition operations. Just by increasing from 2D to 3D, the same sized (3x3x3) kernel convolution will require 6434856 (62x62x62x27) multiplications and 238328 (62x62x62) addition operations. Thus, just adding an extra dimension will increase the computation time and computational complexity by orders of magnitude making the training process impractical. Hence the scaling from 2D to 3D was be done in a way to keep the computation time and complexity within practical limits of the hardware.

The most number of convolution operations (multiplications and additions) are performed in the earlier stages of the CNN when the input image size is the largest. Thus, initial layers were optimized the most for 3D counterparts to keep the computation time and complexity within practical limits. Moreover, a direct reduction in computational complexity is achieved by reducing the channel depth at each layer. Modification of the pooling window and padding size also reduced the computation time and memory requirement of the overall network.

The size of the padding and pooling was also crucial in determining the input size of the successive layers. The input size of the CNNs were also modified to make it suitable to handle 3D anatomical brain data from any MRI scanner. A common preprocessing step in neuroimaging analysis involving multiple participants is normalization to a global atlas. Normalization allows data from different sites and participants to be in a uniform spatial dimension. The input size of the CNNs were thus selected to be the size of the standard MNI atlas which will allow the networks to be used on practically any normalized anatomical MRI data.

Another common feature of the 2D CNN used for images is the color channels for the input layers. Color images consists of red, green and blue channels but neuroimaging data only consists of intensity information. However, brain volumes can be decomposed into various tissue types by the process of segmentation. Segmentation is yet another common preprocessing technique used with neuroimaging data. The three main tissue types in the brain are grey matter (GM), white matter (WM) and cerebrospinal fluid (CSF). Each of the three have unique properties and thus the segmentation maps corresponding to GM, WM and CSF are used as channels in the 3D CNN. AD is a neurodegenerative disease where there is a loss of neuron structure which inturn causes direct changes in the GM and CSF volumes [Frisoni et al., 2002; Wenk, 2003; De Leon et al., 2004]. Thus, having GM, WM and CSF volumes as channels may result in the CNN learning clinically meaningful features.

The memory requirements were also optimized for the 3D CNN. The memory required to store a CNN depends on the total number of parameters. The total number of learnable parameters for 2D AlexNet, GoogleNet and VGG-16 is 61M (million), 7M and 138M respectively while the 3D version has 16.8M, 11.1M and 46.2M respectively. If the CNN was simply scaled up from 2D to 3D, the total number of learning parameters would have increased significantly. However, due to reduction in number of channels and modifications in kernel size, pooling window and padding size, the total number of learnable parameters in the 3D CNN were less than (AlexNet and VGG16) or comparable (GoogleNet) in number to their 2D counterparts. Thus, the memory requirement for storing the 3D CNN would be similar to that of a standard 2D CNN.

The architecture for all 3 CNNs is shown in Figure 2 and Figure 3. Figure 2 shows the architecture for 3D Alexnet and 3D VGG 16 CNN while Figure 3 shows the 3D GoogleNet architecture along with the details of the Inception subnetwork. Different colors indicate different blocks of the CNN while the arrows indicate the dataflow direction. The size of the convolutional filters is denoted for each region with the number of channels indicated within the parenthesis.

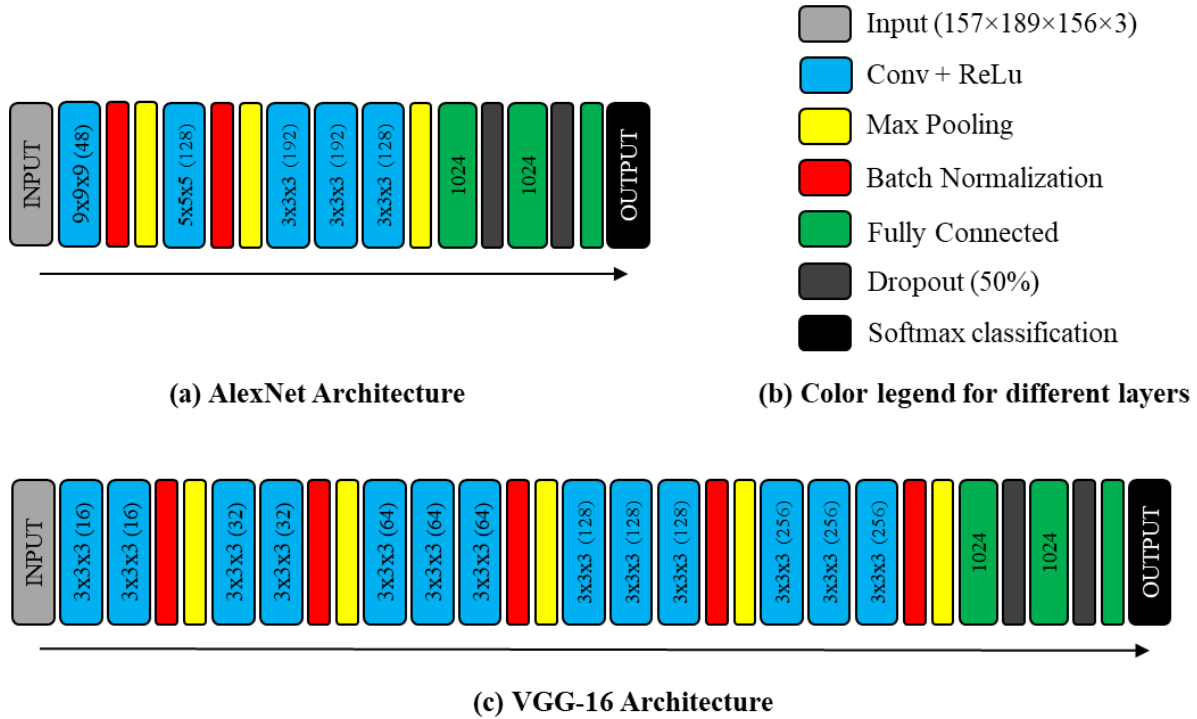

*Figure 2: (a) Architecture of the 3D AlexNet CNN. (b) Color legends for different layers depicted in the figure. (c) Architecture of the 3D VGG16 CNN. [The size of the convolutional filters and the number of filters (in parenthesis) for each of the convolution layer is denoted within each layer. The number of nodes for all the fully connected layers is denoted within the layer except for the last layer which has 3 nodes corresponding to the 3 class. All the max pooling layers were of size 3x3x3 with a stride of 2x2x2.]*

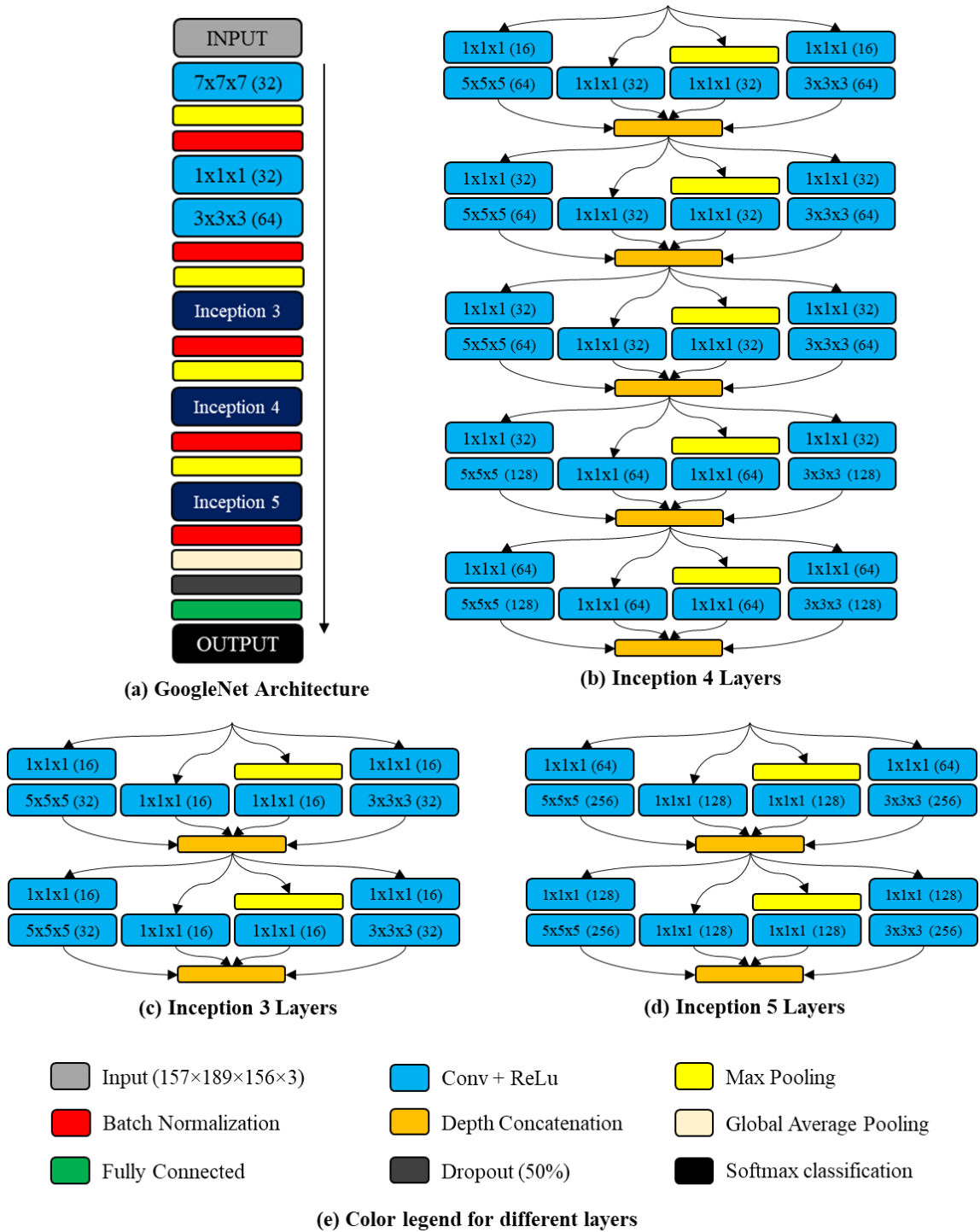

Figure 3: (a) Architecture of the 3D GoogleNet CNN. (b), (c), (d) Architecture of the Inception module at Layer 3, 4 and 5 respectively. (e) Color legends for different layers depicted in the figure. [The size of the convolutional filters and the number of filters (in parenthesis) for each of the convolution layer is denoted within each layer. The number of nodes for the fully connected layers is 3 corresponding to the 3 class. The max pooling in the GoogleNet layers were of size 3x3x3 with a stride of 2x2x2, while within the inception layers was of size 3x3x3 with stride of 1x1x1.]

Alexnet is one of the earliest and one of the simplest CNN. Like the original architecture, the 3D version has 5 convolutional and 3 fully connected layers. The VGG16 architecture is 'deeper' than Alexnet. The 3D VGG16 architecture has 16 learnable layers just like the original one. The uniqueness of VGG architecture is the blocks of cascaded convolutional layers. The uniqueness of GoogleNet is the Inception subnetwork. The inception subnetwork has parallel learnable layers which extracts features at different scales. Unlike the 2D GoogleNet, output is only computed at the end of the last learnable layer and not after every inception subnetwork.

*Training*

The entire dataset was randomized and divided into three parts – training (70%: 1052 samples), validation (15%: 225 samples) and testing (15%: 225 samples). The same split was used for all the three CNNs to be able to have a direct comparison of the performance once trained. All three CNNs were trained using MATLAB on the university's high performance computing center facility. All CNNs were trained using the 'adam' optimizer [Kingma & Ba, 2014]. The networks were trained for a total of 1024 epochs. The initial learning rate was set to $1 \times 10^{-5}$ (1e-5), with learning rate reducing by a factor of 0.75 after every 256 epochs. The training was performed in minibatch of size 32 with all training samples being randomized after every epoch. The validation frequency was set to 128 iterations which is roughly equivalent to 4 epochs. To avoid overfitting, L2 regularization was used with a regularization constant ($\lambda$) of 0.1. Overfitting was also reduced by using dropout layers after every fully connected layer for all 3 CNNs. The dropout rate was set to 0.5. A 5-fold cross validation was also performed for each of the three CNN.

*Analysis*

Once trained, performance parameters were computed for all three CNNs using the confusion matrix (for validation and testing partitions). The confusion matrices were used to calculate the class-wise accuracy, precision, sensitivity, specificity and F1 score. Class-wise ROC (Receiver Operating Characteristic) curves were also generated, and corresponding AUC (Area Under the Curve) was calculated. The performance parameters were also computed for each of the 5 folds of the 5-fold cross validation. Because the same data split was used for training of each of the three CNNs, the predictions from the three CNNs were combined using an ensemble algorithm. The ensemble algorithms were applied to data samples not used for training, i.e., combined validation and testing split. Ensemble averaging and voting ensemble algorithms were compared. In ensemble averaging, the class-wise prediction probabilities were averaged to obtain ensemble probabilities for each class and the input sample was assigned to the class with highest ensemble probability. In voting ensemble, the input sample was assigned to the class with the highest number of prediction votes. Because there are 3 class and 3 separate CNN models, it is possible that each class may get a single vote. In that case, the input will be assigned to the class with highest prediction probability among all three CNNs.

# RESULTS

The summary of the classification performance for all the three CNN is shown in Figure 4. The left and the right part of the figure shows class wise ROC curve and the corresponding confusion matrix for validation and testing data split respectively. The middle portion shows the plot of class wise performance parameters for both validation and testing data split. For the confusion matrix, the horizontal axis represents the true class while the vertical axis represents predicted class. From the figure it can be observed that all the classifiers are performing well in distinguishing between different classes. The best classification performance is observed for AD class followed by CN and then MCI.

The results of the 5-fold cross validation are shown in Figure 5. Different subplots in the figure corresponds to different performance parameters. Each subplot shows a boxplot of the class wise performance parameter. The overall high values of accuracies indicates that each of the CNN architecture is capable of performing the classification task in a satisfactory manner. From the figure it can be observed that the Alexnet and GoogleNet architectures are performing slightly better than the VGG16 architecture. As noted in Figure 4, it can be observed again in Figure 5 that MCI is the most challenging class to be classified correctly.

The performance of the ensemble algorithm is shown in Figure 6. In the figure, performance parameters of both the ensemble approach are compared with individual CNN architectures. As expected, the ensemble approach performs better than individual CNN. Moreover, the performance of both the ensemble approach is almost similar.

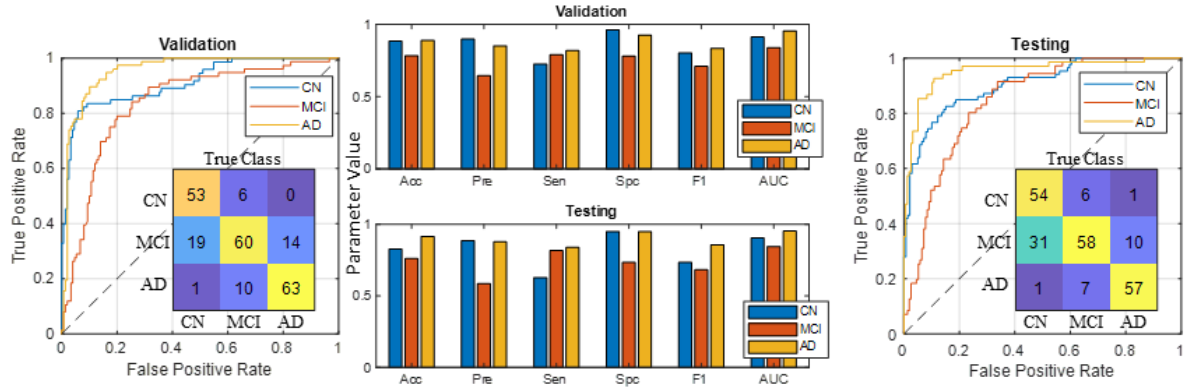
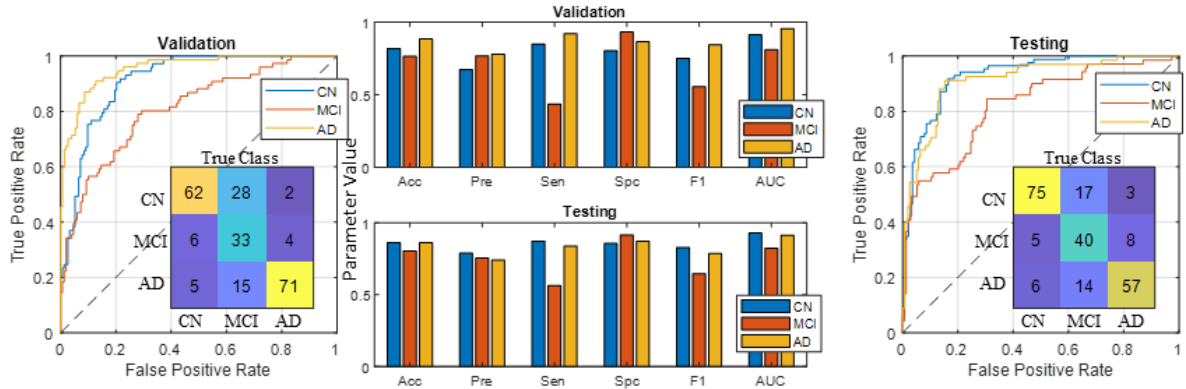
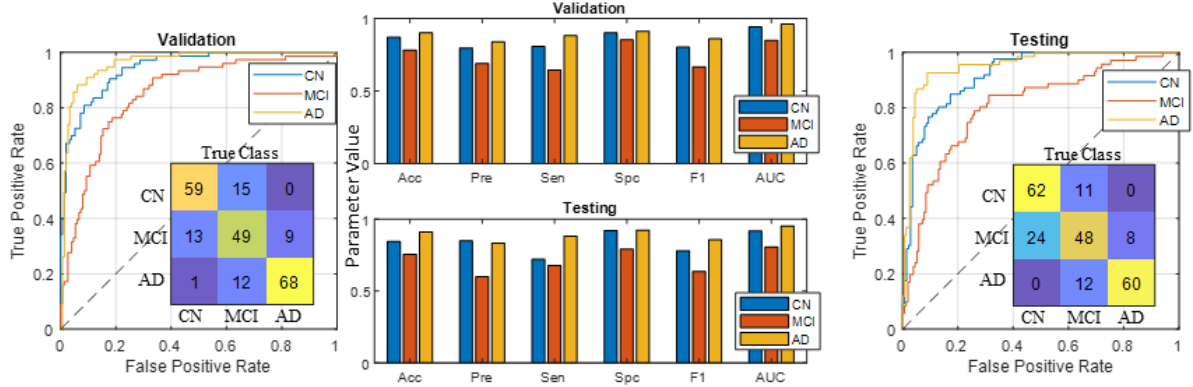

*Figure 4: Performance parameters of validation and testing data split for all 3 CNN architecture (rows). The plots on the left and right are the class-wise ROC curve for validation and testing data split. The confusion matrix is also shown on the subplot The colors correspond to the value in the cell. Larger numbers are given lighter color (yellow) while smaller numbers are given darker color (blue). The bar chart in the center shows class-wise network performance parameters (Acc: accuracy, Pre: precision, Sen: sensitivity, Spc: specificity, F1: F1-score, and AUC: area under the curve).*

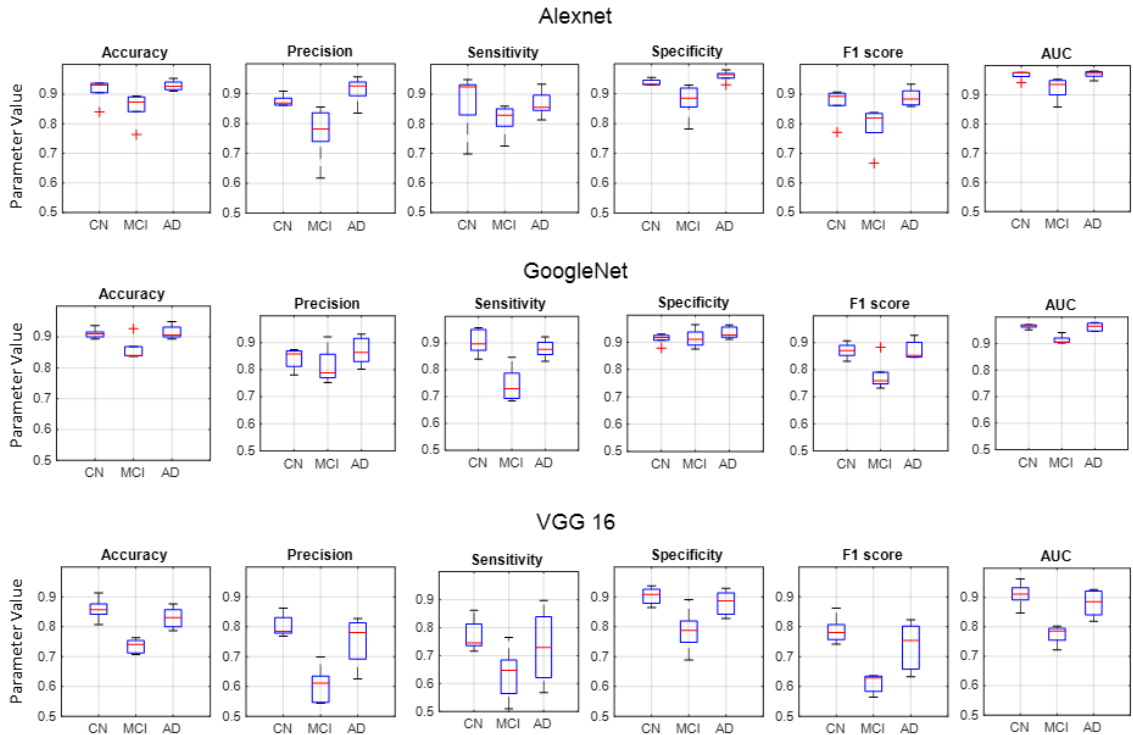

*Figure 5: Boxplots displaying the 5-fold cross validation results for (top) AlexNet, (middle) GoogleNet and (bottom) VGG16 architectures. The x-axis represents true class while the y-axis represents parameter value.*

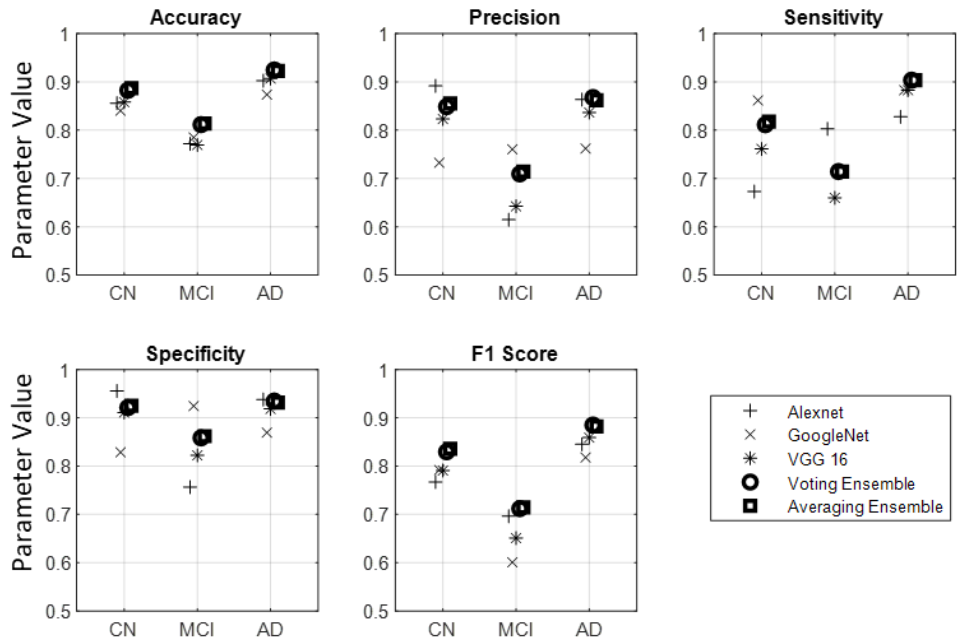

*Figure 6: Performance comparison of individual CNNs and ensemble algorithm. The x-axis represents true class while the y-axis represents parameter value.*

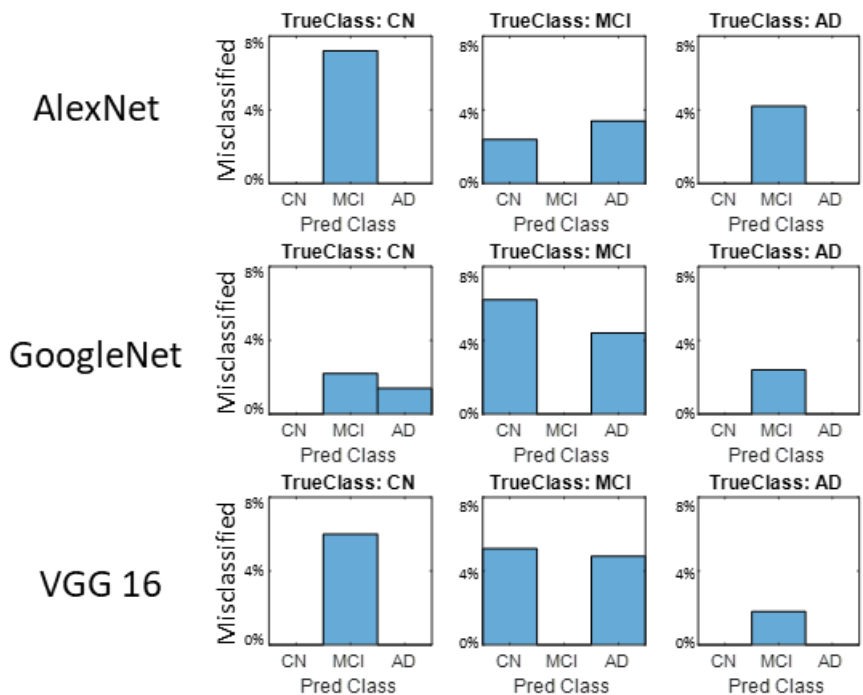

*Figure 7: Class-wise histogram for the misclassified samples for all 3 CNN architecture. Each column subplot represents a different class while each row subplot represents different CNN architecture. The x-axis shows the predicted class (Pred class) while the y-axis shows the percentage misclassified for a given true class.*

## DISCUSSION

### Misclassification analysis

For any given input, each of the classifier outputs a probability indicating the likelihood of the input being in a particular class. The input is assigned to the output type with the highest probability. For the samples classified correctly, the highest probability is given to the correct category. However, for misclassified samples, the second highest priority or the third highest priority class is the actual correct category. For all the misclassified samples, a misclassification analysis was conducted. The histogram of the highest probability class was generated for all the misclassified samples in all 3 CNN architectures. Figure 7 shows the class wise (columns) misclassification histogram for all 3 CNN architecture (rows). For true class AD, it can be observed that all the misclassification is predicted as MCI class while for all the true class CN except Googlenet, the misclassifications are predicted as MCI. Given the fact that 'sharp demarcations between normal cognition and MCI and between MCI and dementia are difficult, and clinical judgment must be used to make these distinctions.' [Albert et al., 2011, p 271] it is indeed a good sign that the networks are not misclassifying true class AD as CN and vice versa. True class MCI is being misclassified as both AD and CN and given the fact that MCI is an intermediate stage between both AD and CN it is challenging to have a hard demarcation between these classes. All these misclassification results suggests that the deep learning networks are actually making predictions which are sensible and not just random. Such misclassification events can be fancily called 'task failed successfully.'

**Feature Extraction**

Features that were weighted mode by the deep learning networks were analyzed. Saliency maps were generated for each network and features were extracted for each class [Simonyan et al., 2013; Zeiler et al., 2014]. Figure 8 shows the saliency map of important features for each of the three network and each category. One important observation is that all three networks have some overlap with the hippocampus region and consider that as an important feature for classification of AD. Saliency maps for AlexNet and VGG16 overlap largely with the right hippocampi while that of GoogleNet overlaps in both the hemispheres. Hippocampus has been shown to be a key brain region of interest for detection and classigication of Alzheimer's Disease [Laakso et al., 1998; Laakso et al., 2000; Rössler et al., 2002; Wang et al., 2006; Rao et al., 2022]. All the three CNNs giving more importance to the hippocampus regions indicate that the networks are actually extracting useful features to make the decision about the category.

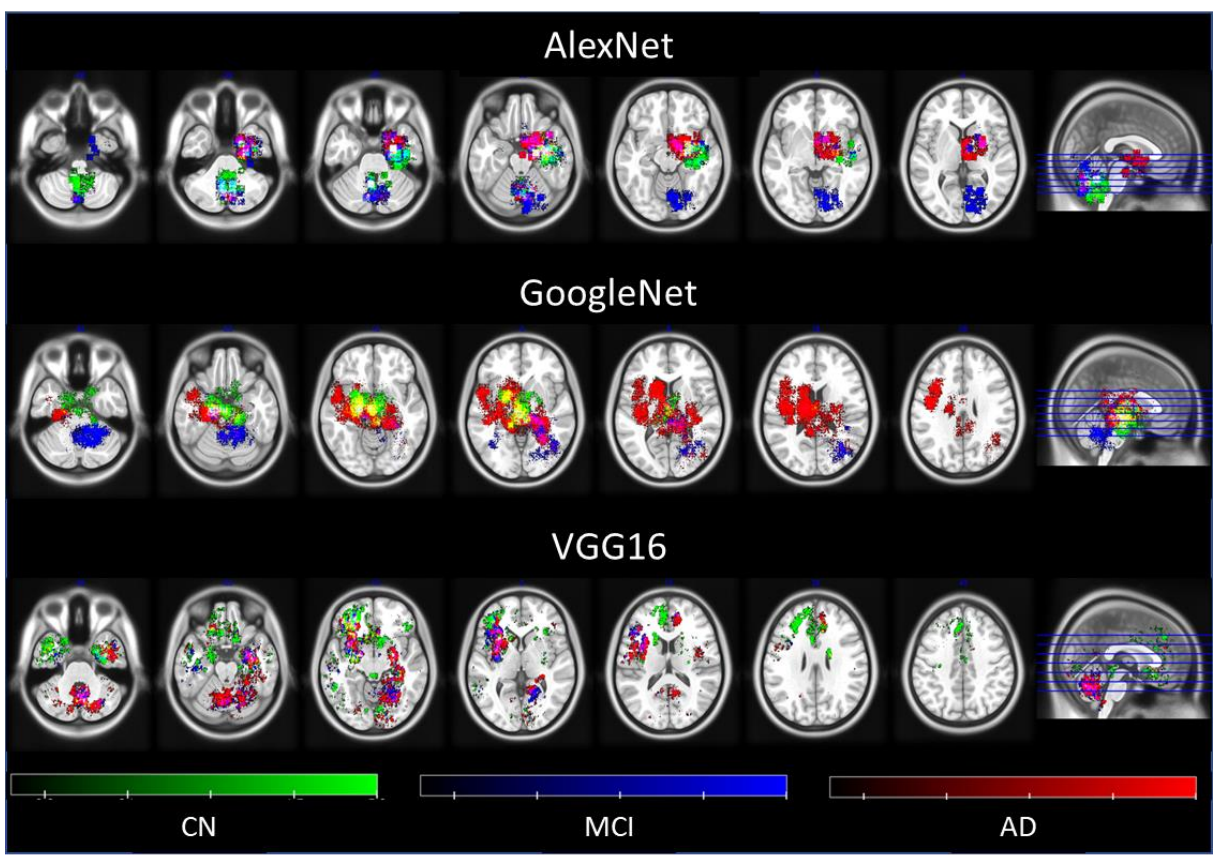

*Figure 8: Saliency map for (top) AlexNet, (middle) GoogleNet and (bottom) VGG16 architectures mapped onto MNI brain atlas. Maps for different categories are represented in different colors. [The strength of the color depends on the strength of the output of the CNN features (arbitrary value). The color coding is additive in nature.]*

For GoogleNet, the saliency maps also include parts of the caudate nucleus for AD and CN categories. The volumetric analysis of the caudate nucleus has shown significant difference between AD and CN categories in previous studies [Barber et al., 2002; Jiji et al., 2013]. It has also been shown that the absolute caudate nucleus volume and the ratio of caudate nucleus to hippocampus volume increases for Alzheimer's Disease [Persson et al., 2018]. An interesting observation is the overlap of saliency maps with the cerebellum, mainly for the MCI category. Some studies do suggest degenerative changes of the cerebellum during MCI [Thomann et al., 2008] and Alzheimer's Disease dementia in general [Jacobs et al., 2018]. Again, observing meaningful anatomical features in the saliency maps indicate that the networks are doing a decent job of feature extraction and some of the features can be post processed and used by human experts to make further inferences.

**Future Work**

The results from the current study suggest that full brain anatomical MRI volumes can be used along with existing deep learning algorithms to make predictions about the stages of Alzheimer's Disease dementia. The ensemble algorithm performs better than any single CNN and the extracted features by each of the networks have some meaningful anatomical information. The future work for this study is twofold. First, train the models with a larger dataset and have a provision for constant update of the CNNs i.e., continuous online training with new input data. The network can be trained to differentiate between more categories like early and late MCI (EMCI and LMCI) or MCI due to Alzheimer. Second, development of a software tool that can generate predictions and output feature maps which can then be used by a radiologist/expert to assist in better decision making. Instead of a hard classification, the software can output probabilities of various categories and then the human expert can make an educated decision. For instance, for a given brain volume, the CNN can output the probability of various predicted class along with the saliency maps and then a radiologist can examine both raw and processed data an make the final decision. Because the networks only need to be trained once, the post processing/prediction with new data should be fast (in terms of computation time) and direct (in terms of computational complexity).

# CONCLUSION

In this study, it has been successfully shown that different kinds of deep learning algorithms can be used to predict/classify various stages of Alzheimer's Disease dementia using standard anatomical MRI brain images (3D volume). Well known CNNs used for 2D images can be modified and used with 3D brain volumes to detect useful neurological information. The ability to detect intermediate stages of AD like MCI may be useful for preventing or delaying the onset of AD. Such an early detection ability may prove to be very beneficial for the patient. One of the main features of the 3D CNN discussed here is the simplicity on the user's end. Once trained, the input to these CNN is full brain volume which can be obtained directly from the MRI scanner.

An inbuilt preprocessing and trained network can be then deployed as a software package for the radiologists and physicians who may not be very proficient with the nitty gritty of computer programming. The performance of the ensemble algorithm is more promising than any individual CNN and the features (saliency maps) extracted by these CNNs can provide additional information to the radiologists/physicians. A human deep learning synergy can be beneficial in diagnosis of neurodegenerative disorders like Alzheimer's Disease.

*Data Availability Statement*

The datasets generated for this study and the MATLAB codes used for the analysis of the data are available on request to the corresponding author. Requests to access these datasets should be directed to HP, harshit.parmar@ttu.edu.